\documentclass[letterpaper, 10 pt, conference]{ieeeconf}  

\IEEEoverridecommandlockouts                              
\let\labelindent\relax
\usepackage{enumitem}
\usepackage[bookmarks=true]{hyperref}
\usepackage{tabularx}
\usepackage{graphicx}
\usepackage{lipsum}
\usepackage{epsfig}
\usepackage{array}
\usepackage{subfig}
\usepackage{cite}
\usepackage{float}
\usepackage{ifthen}
\usepackage{xcolor}
\usepackage{ntheorem}
\usepackage{algpseudocode}
\usepackage{algorithm}
\usepackage{booktabs} %
\usepackage{array}    %
\theoremseparator{:}

\usepackage[font=footnotesize]{caption}

\newcommand{\fig}[1]{Fig.~\ref{#1}}

\setlist[itemize]{leftmargin=*}

\title{\LARGE \bf Human-Robot Collaboration for the Remote Control of Mobile Humanoid Robots with Torso-Arm Coordination}

\author{Nikita Boguslavskii$^{1}$, Lorena Maria Genua$^{1}$ and Zhi Li$^{1}$
\thanks{$^{1}$ Robotics Engineering Department, Worcester Polytechnic Institute (WPI), Worcester, MA 01609, USA {\tt\small \{nbboguslavskii,lgenua,zli11\}@wpi.edu}}
}

\begin{document}

\onecolumn
\thispagestyle{empty} %

{\LARGE \textbf{IEEE Copyright Notice}} %

\vspace{1.5em} %

{\large
© 2025 IEEE. Personal use of this material is permitted. Permission from IEEE must be obtained for all other uses, in any current or future media, including reprinting/republishing this material for advertising or promotional purposes, creating new collective works, for resale or redistribution to servers or lists, or reuse of any copyrighted component of this work in other works.
}

\newpage
\twocolumn

\maketitle

\begin{abstract}
Recently, many humanoid robots have been increasingly deployed in various facilities, including hospitals and assisted living environments, where they are often remotely controlled by human operators.
Their kinematic redundancy enhances reachability and manipulability, enabling them to navigate complex, cluttered environments and perform a wide range of tasks. 
However, this redundancy also presents significant control challenges, particularly in coordinating the movements of the robot's macro-micro structure (torso and arms).
Therefore, we propose various human-robot collaborative (HRC) methods for coordinating the torso and arm of remotely controlled mobile humanoid robots, aiming to balance autonomy and human input to enhance system efficiency and task execution.
The proposed methods include human-initiated approaches, where users manually control torso movements, and robot-initiated approaches, which autonomously coordinate torso and arm based on factors such as reachability, task goal, or inferred human intent.
We conducted a user study with N=17 participants to compare the proposed approaches in terms of task performance, manipulability, and energy efficiency, and analyzed which methods were preferred by participants.

\end{abstract}

\section{Introduction}\label{sec:intro}

Human-robot collaborative (HRC) control enables humans and robot autonomy to complement each other and improve overall robotic manipulation performance. 
In general, robot autonomy can be utilized to control precise and structured manipulation, and generate optimal motion plans if some optimization objectives (e.g., energy, manipulability) are clearly specified. 
Meanwhile, humans can leverage their task knowledge and experience to provide robots with helpful guidance and corrections to task and motion plans, considering complex trade-offs---which could be time-varying, contextual, and difficult to encode in cost/reward function. 
In this paper, we focus on robotic manipulators with a moving torso, such as the mobile humanoid robots commonly used for nursing and living assistance (e.g.,~\cite{wise2016fetch,ackerman2018moxi,marques2022commodity,boguslavskii2023shared}). 

While the kinematic redundancy resulting from the torso-arm coordination can improve the robot’s reachability and manipulability in a cluttered environment, it also introduces additional control complexity and considerations for a desired human-robot collaboration. 
To this end, we propose and compare several human-robot collaborative control approaches for the macro-micro structure coordination of robotic manipulators, balancing robot performance, human factors, and task requirements. 

For torso-arm coordination, our proposed approaches include \textit{human-initiated} methods, which allow users to control both continuous and discrete torso movements, and \textit{robot-initiated} methods, which autonomously coordinate torso and arm motions based on factors such as reachability, manipulability, energy efficiency, and human intent.

In a user study with 17 participants, we compared our HRC methods against \textit{RelaxedIK}~\cite{rakita2018relaxedik}, a state-of-the-art Inverse Kinematics (IK) solver designed for redundant manipulators. We identified the most effective control modes by analyzing user performance, preferences, and expertise, gaining insights into which approaches suit different experience levels. These findings contribute to research on human robot collaboration by introducing and evaluating different control strategies for torso-arm coordination.

\section{Related Work}\label{sec:related}
\begin{figure*}[t]
    \centering
    \includegraphics[width=0.98\textwidth]{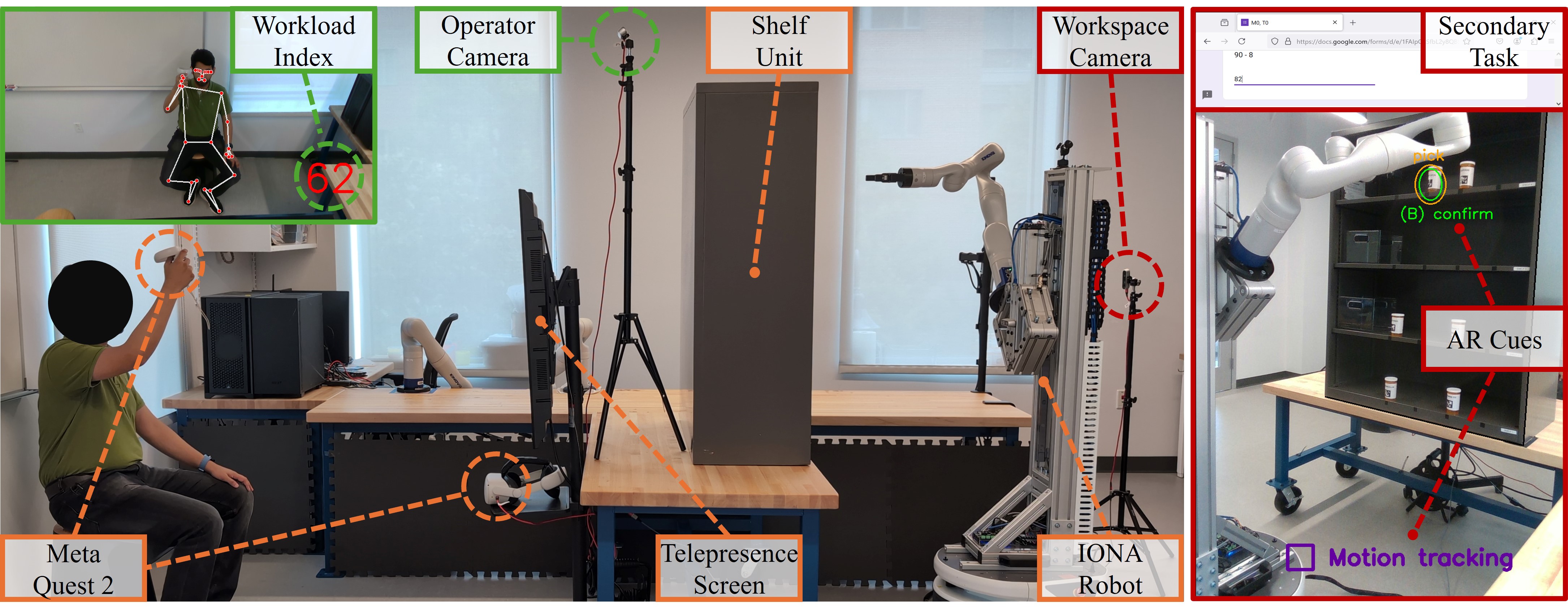}
    \caption{The experimental setup consists of two workspaces. The robotic workspace features a shelf unit with four shelves. IONA robot featuring a motorized torso and a 7-degree-of-freedom robotic arm is located next to the shelf. A workspace camera captures the torso, arm and provides a complete view of the manipulation area. The operator workspace includes a screen displaying a graphical user interface (GUI), which streams video from the workspace camera with augmented reality (AR) cues overlaid. The operator uses a MetaQuest 2 handheld controller to control the robot. An additional operator camera is placed above the screen to capture the participant's posture, with key body landmarks tracked for physical workload estimation.}
    \label{fig:002System}
    \vspace{-3ex}
\end{figure*}
Thus far, the research on HRC control for remote robot manipulation mostly focuses on the shared autonomy for controlling the robot’s end-effector motions (e.g.,~\cite{dragan2013policy,javdani2018shared,nikolaidis2017human,hagenow2021corrective,thakur2024mocap,jonnavittula2024sari}).  
In some recent work, more advanced HRC approaches have been proposed such that robot autonomy can assist the human in different aspects of the manipulation tasks, from perception, to decision-making, to action and motion control. 
The robot can also leverage its task, action and motion planning to assist humans with different levels of autonomy~\cite{perez2023experimental,lin2023impacts,krishnan2023human}. 
However, existing shared autonomy approaches haven’t been extended to the torso-arm coordination of (mobile) humanoid robots. 
On the other hand, in motion planning, the torso-arm coordination problem has been typically addressed using the generic approaches for kinematic redundancy resolution~\cite{haviland2022holistic}, considering all the DoFs from the base and arm that can affect the robot end-effector pose. Alternatively, it can also be addressed using separate motion planners for the macro- and micro-structure and introduce a policy for their coordination~\cite{yokoyama2023asc}. Because most of the manipulation tasks prioritize the end-effector’s operation, the planning of the micro-structure is prioritized and will determine the constraints for planning the motion of the macro-structure (e.g., ~\cite{reister2022combining}). 
In either case, the behavior of the autonomous torso-arm coordination can be customized if humans can clearly specify the constraints and optimization objectives for the motion planner. 

However, in real-time operation humans have little opportunities to intervene in the torso-arm coordination, particularly if they need to tradeoff multiple and contextual considerations for the robot, task and human performance, which could be hard to encode in some reward or cost function. 
To address these limitations in the related work, we contribute several human-robot collaborative control approaches for torso-arm coordination and provide experimental evaluation for their effectiveness. Instead of declaring a single optimal solution, we provide a collection of our methods to be selected according to the robot system, task and types of operators.

\section{Proposed HRC Torso-Arm Coordination}\label{sec:method}

Our proposed methods consider factors affecting human ability and experience in collaborative manipulation (e.g. improving the consistency and predictability of robot autonomy, avoiding visual occlusion, and reducing human's operation workload), in addition to what may affect the robot and task performance (e.g., the reachability and manipulability of the end effector, the robot operation speed, and the energy consumption). 
We propose \textit{human-initiated} approaches, which allow humans to manage torso-arm coordination, and the \textit{robot-initiated} approaches, which delegate control of robot's torso to robot autonomy.

\subsection{Human-Initiated Approaches}\label{sec:method-torso+arm-human}

\begin{algorithm}
\caption{Preset Heights (PH)}\label{alg:preset_heights}
\begin{algorithmic}

\Function{PresetHeightsControl}{}
    \State preset\_heights $\gets$ [BOTTOM, MIDDLE, TOP]
    \State pos\_ix $\gets$ FindCurrentPositionIndex()
    
    \State input $\gets$ GetJoystickInput()
    
    \If{input = UP \textbf{and} pos\_ix $<$ length(preset\_heights) - 1}
        \State pos\_ix $\gets$ pos\_ix + 1
    \ElsIf{input = DOWN \textbf{and} pos\_ix $>$ 0}
        \State pos\_ix $\gets$ pos\_ix - 1
    \EndIf
    
    \State MoveTorsoTo(preset\_heights[pos\_ix])
\EndFunction
\end{algorithmic}
\end{algorithm}
\vspace{-1ex}

During the human-initiated coordination, the user is responsible for the full management of the interaction between the robotic arm and the torso, i.e. adjusting positions of both components independently to perform the task. 
We propose two human-initiated approaches that allow experienced operators to have more flexibility to control the torso-arm coordination and increase their level of engagement. 

\textbf{Velocity Torso Control (V)} mode involves controlling the torso using a joystick on the handheld controller, while the robotic arm is independently controlled using the operator's hand movement. This mode offers the most granular control over the torso's position in a continuous space. We have selected it as our human-initiated baseline.

The \textbf{Preset Heights (PH)} control mode segments the workspace into several regions of interest (e.g., "top", "middle", and "bottom") and allows the user to transition between them by pressing the controller's joystick up or down, while the arm is controlled independently with the controller. The coordination remains manual, but restricting the torso motion to a set of predefined positions reduces the number of choices for the user, making the selection and adjustment to a desired torso pose faster.

\subsection{Robot-Initiated Approaches}\label{sec:method-robot}

Our robot-initiated approaches are designed for less experienced operators who may prefer robot autonomy to manage torso-arm coordination.
These methods enable operators to focus on controlling micro-structure manipulation, while the robot autonomously handles macro-structure movements to optimize task performance. 
\textit{Reactive autonomy} methods adjust the macro-structure in response to a specific trigger, such as user input, system state, or external factors. The adjustment occurs at the moment the trigger is detected. 
The \textbf{Proximity (P)} mode moves the robot torso only when the end-effector approaches its motion range limits in the vertical space, aiming to reduce energy consumption for moving the macro-structure. The intuition behind this mode is that the torso is adjusted in a discrete manner only when necessary to accommodate the reachability of the robotic arm. As a result, the torso motor remains mostly inactive, leading to lower energy consumption. However, this mode does not guarantee optimal manipulability.
To prevent the robot's autonomous torso motion from interfering with the operator's control of the end-effector, the Proximity (P) mode includes a compensation feature. When the torso moves up, the robot compensates by lowering the end-effector by the same distance, allowing it to maintain a consistent height within the workspace.

\begin{algorithm}
\caption{Proximity Method (P)}\label{alg:proximity}
\begin{algorithmic}

\Function{ProximityMethod}{}
    \State arm\_z\_pos $\gets$ GetArmZPosition()
    \State torso\_pos $\gets$ GetTorsoPosition()

    \State upper\_limit $\gets$ ARM\_UPPER\_BOUNDARY
    \State lower\_limit $\gets$ ARM\_LOWER\_BOUNDARY

    \If{arm\_z\_pos $\geq$ upper\_limit}
        \State delta\_pos $\gets$ arm\_z\_pos - upper\_limit
        \State torso\_pos $\gets$ torso\_pos + delta\_pos
        \State MoveTorso(torso\_pos)
        \State \Call{Compensation}{delta\_pos}
    \ElsIf{arm\_z\_pos $\leq$ lower\_limit}
        \State delta\_pos $\gets$ lower\_limit - arm\_z\_pos
        \State torso\_pos $\gets$ torso\_pos - delta\_pos
        \State MoveTorso(torso\_pos)
        \State \Call{Compensation}{delta\_pos}
    \EndIf
\EndFunction

\end{algorithmic}
\end{algorithm}

\begin{algorithm}
\caption{Torso-Arm Compensation}\label{alg:compensation}
\begin{algorithmic}

\Function{Compensation}{delta\_pos}
    \State arm\_z\_pos $\gets$ GetArmZPosition()
    \State arm\_z\_pos $\gets$ arm\_z\_pos - delta\_pos
    \State MoveArmTo(arm\_z\_pos) 
\EndFunction

\end{algorithmic}
\end{algorithm}

The \textbf{Scaling (S)} mode scales (maps) the user's arm workspace to the combined vertical workspace of both the robotic arm and the torso, enabling better and faster manipulation by covering the robot's entire vertical range with human arm movements. As a result, human hand motion is scaled up. However, it is less energy efficient because of its continuous motion style, which keeps the torso motor active, leading to higher energy consumption.

\begin{algorithm}
\caption{Scaling Method (S)}\label{alg:scaling}
\begin{algorithmic}

\Function{ScalingMethod}{}
    \State human\_arm\_pos $\gets$ GetHumanArmPosition()
    \State human\_ws $\gets$ GetHumanWorkspace()

    \State arm\_ws $\gets$ GetRobotArmWorkspace()
    \State torso\_ws $\gets$ GetTorsoWorkspace()

    \State torso\_scale $\gets$ torso\_ws / human\_ws
    \State arm\_scale $\gets$ robot\_arm\_ws / human\_ws

    \State target\_torso\_pos $\gets$ human\_arm\_pos $\times$ torso\_scale
    \State target\_arm\_pos $\gets$ human\_arm\_pos $\times$ arm\_scale

    \State \Call{MoveTorso}{target\_torso\_pos}
    \State \Call{MoveRobotArm}{target\_arm\_pos}
\EndFunction

\end{algorithmic}
\end{algorithm}

In the \textbf{Chasing (C)} mode, the robot torso continuously follows the robot arm's vertical motion to maintain optimal manipulability and reachability. Unlike the Scaling mode, however, it maintains a 1:1 mapping between human hand motion and the system. Like Proximity mode, it also features vertical compensation.

\begin{algorithm}
\caption{Chasing Method (C)}\label{alg:chasing}
\begin{algorithmic}

\Function{ChasingMethod}{arm\_z\_pos\_prev}
    \State torso\_pos $\gets$ GetTorsoPosition()

    \State arm\_z\_pos $\gets$ GetArmZPosition()
    \State delta $\gets$ arm\_z\_pos - arm\_z\_pos\_prev

    \If{delta $\neq$ 0}
        \State torso\_z\_pos $\gets$ torso\_pos + delta
        \State \Call{MoveTorso}{torso\_z\_pos}
        \State \Call{Compensation}{delta}
    \EndIf

    \State arm\_z\_pos\_prev $\gets$ arm\_z\_pos
\EndFunction

\end{algorithmic}
\end{algorithm}

\textit{Proactive autonomy} anticipates the end-effector position requirements and adjusts the torso in advance. Rather than waiting for a trigger, these methods predict the next necessary torso position based on the user’s actions, intent, or task sequence. By adjusting ahead of time, proactive autonomy minimizes delays in torso movement, enabling smoother coordination and improving efficiency, especially in tasks that require rapid or precise movements.

\textbf{Task-based (TB)} method anticipates the user's future movements based on the task state and sequence, adjusting the torso in advance to optimize end-effector reachability and manipulability. Its discrete motion style and optimal torso position prediction enhance energy efficiency. It also features the vertical compensation algorithm, like Proximity and Chasing. In our current implementation, an expert predefined the torso position sequence based on the task requirements.

\begin{algorithm}
\caption{Task-Based Method}\label{alg:task_based}
\begin{algorithmic}

\Function{TaskBasedMethod}{task\_state}
    \State torso\_pos $\gets$ GetTorsoPosition()
    \State predicted\_torso\_pos $\gets$ PredictTorsoPosition(task\_state)

    \State delta $\gets$ predicted\_torso\_pos - torso\_pos
    \State \Call{MoveTorso}{predicted\_torso\_pos}
    \State \Call{Compensation}{delta}
    
\EndFunction

\end{algorithmic}
\end{algorithm}

As a baseline approach for robot-initiated control, our \textbf{RelaxedIK (RIK)} mode adapts the state-of-the-art RelaxedIK solver~\cite{rakita2018relaxedik} to resolve the inverse kinematics for the integrated torso-arm system. The RelaxedIK optimizes for manipulability, reachability, and self-collision avoidance, and motion smoothness, but does not account for the energy consumption and slow motion caused by frequent torso motions. In our implementation, no additional weight tuning was performed.

For robotic arm control in all other modes, we used the RelaxedIK model, which does not incorporate the torso's prismatic joint and was set up for the Kinova Gen3 7-DoF in our previous work~\cite{boguslavskii2023shared}.

\section{User Study Evaluation}\label{sec:exp}

\subsection{Experiment setup}

We conducted a user study with N=17 participants to compare the proposed HRC torso-arm coordination approaches. 
Shown in~\fig{fig:002System}, the human operator and robot are located in separate workspaces. 
The robot workspace contains a four-level shelf unit with medicine containers to be organized.
In the human workspace, the operator uses a screen-based graphical user interface (GUI) streaming images from the robot workspace camera.
An RGB+D camera is set up to track the operator's body posture in real time and estimate their physical workload~\cite{lin2023impacts, lin2024freeform} during the task.

\subsection{Task Design}

The simulated task required the operator to control the robot to rearrange medicine containers across the shelves, following a predefined order. The shelf height requires coordination between the torso and robotic arm to reach all levels effectively.

The task involves both \textbf{short-range} operations, where objects are positioned close to each other, requiring little or no torso movement, and \textbf{long-range} operations, where the torso must make significant adjustments for reachability. The task consists of 12 sub-tasks, comprising a balanced mix of short- and long-range object picking and placing. This design ensures a comprehensive evaluation of macro-micro torso-arm coordination.

To ensure consistency across participants, the task procedure was strictly controlled. Operators followed a predefined sequence of pick-and-place operations, guided by augmented reality (AR) cues. The robot's workspace camera provided real-time feedback, highlighting target objects with an orange circle to aid in selection. Once the robot's end-effector was close to the target, the operator followed the AR instructions and pressed a button to confirm the target for pick-and-place. The robot then autonomously performed the precise action. 
In cases where objects were beyond the robot's reach, an "Out of Reach" prompt appeared, signaling the operator to move the torso.

Additionally, to assess cognitive workload, participants were required to verbally solve math problems displayed in the graphical user interface (GUI) while controlling the robot (See \fig{fig:002System}). This secondary task provided an objective measure of mental effort during robotic control.

\subsection{Experiment Procedure}

For each participant, the experiment consists of three sessions. 
In the \textbf{Training Session}, the experimenter verbally explained the experimental setup, including the robot platform, the telepresence interface with camera view with AR cues, the MetaQuest 2 controller and its functions for robot torso and arm motion control. The participants also practiced the robot control until they feel comfortable with the robot operation. It was explicitly stated that the task sequence would remain the same throughout the study to minimize any learning effects related to task knowledge.

Once participants felt comfortable with the controls, they proceeded to the \textbf{Performance Session}. During this phase, they performed two trials of the task using the Velocity Control mode, aiming to complete it as quickly and accurately as possible while minimizing wrong target selections. Task completion time was recorded to assess each participant's entry-level performance, ensuring that initial experience biases were reduced. After completing both trials, participants subjectively reported their performance and confidence in controlling the robotic arm.

In the \textbf{Evaluation Session} participants completed two trials for each of the seven torso-arm coordination modes. The order of the modes was randomized for each participant to minimize ordering effects. Each mode's principle was explained and briefly practiced by participants before starting the trials. After both trials, participants provided brief oral feedback, comparing the mode they had just used to others they had tried and discussing their likes and dislikes. This freeform reflection offered valuable insights and helped participants complete both standard and custom questionnaires more effectively following the interview.

After finishing both trials for all the modes, the participants ranked top-three preferred modes and commented on their overall experience and the best and worst modes they encountered. Each experiment session took approximately 2 to 2.5 hours, and participants were compensated with a \$5 Amazon gift card for every 30 minutes of participation.

\subsection{Participants}\label{sec:result}

We recruited 17 participants for the study, comprising 11 males and 6 females. Their mean age was 25.71 years (\textit{SD = 2.7, Min = 21, Max = 32}), and the mean height was 175 cm (\textit{SD = 8.72}). The participants reported 14 right-hand dominant and 3 left-hand dominant individuals, with 9 having robotics experience and 8 without. Before the study, their VR controller experience averaged 3.12/7 (\textit{SD = 1.91}), and robotic teleoperation experience was 2.41/7 (\textit{SD = 1.68}). After the training phase, subjective confidence in using the robotic arm was reported as 5.65/7 (\textit{SD = 0.68}), and post-performance session, subjective confidence rose to 6.41/7 (\textit{SD = 0.6}), indicating significant improvement (p $<$ 0.001). The average performance task completion time was 105.83 seconds (\textit{SD = 33.34 seconds}).

\subsection{Evaluation Metrics and Measurements}

Our data analysis compares the seven HRC torso-arm using the objective metrics including: 1) the \textbf{task completion time}, for entire tasks and for long- and short-range motion sub-tasks, respectively; 2) the robotic arm \textbf{manipulability index}; 3) \textbf{torso energy efficiency}, assessed by torso motion time (absolute time measure) and the fraction of torso motion in each trial (normalized by trial duration). These metrics assume minimal to no energy consumption when the torso is not in motion, as it remains supported by a magnetic brake. The robotic arm's energy consumption was not estimated; 4) the operator's \textbf{cognitive workload}, assessed by the number of math problems they answered correctly and their response rate (normalized by trial duration); 5) the operator's \textbf{physical workload}, assessed using the index, estimated through real-time tracking of the operator's posture using the method from our prior work in~\cite{lin2023impacts, lin2024freeform}. 

Our subjective evaluation included: 1) the \textbf{NASA-TLX and SUS} surveys; 2) the \textbf{customized questionnaire} in which the participants rated the perception of \textit{robotic arm reactions}, the \textit{ease of robotic torso control}, the completeness of torso control (i.e., feeling of control over the torso), and the \textit{alignment of torso motion with user intent}, on the scale from 1 to 5; 3) the \textbf{top-three preferred modes} reported by participants based on their overall experience. Given the similarities among several modes, this approach helps identify patterns in user preferences, highlighting favored principles or features rather than specific implementations.

\section{Results}\label{sec:result}

The hypotheses for this study were selected based on the specific strengths of each proposed method or groups of methods and aim to compare these methods against each other where applicable, as well as against the baseline and state-of-the-art systems, to evaluate their effectiveness in the teleoperations scenario based on different criteria. 

The collected objective data did not meet the normality assumptions for ANOVA, as confirmed by the Shapiro-Wilk test. Consequently, the analysis was conducted using the Kruskal-Wallis test by ranks, with pairwise comparisons performed using Dunn's test with Bonferroni correction.

\subsection{Overall Task Completion Time}\label{sec:result-time}

\begin{itemize}

\item \textbf{\textit{H1.1.a.}} \textit{V} will be outperformed by all other methods.

\item \textbf{\textit{H1.1.b.}} \textit{S} will outperform Manual methods (\textit{V} and \textit{PH}), all other Reactive Autonomies (\textit{P, C}) and \textit{RIK} method. We hypothesize that \textit{S} method will be overall the fastest due to the scaled motion mapping.

\item \textbf{\textit{H1.1.c.}} \textit{C} will outperform \textit{PH} and \textit{RIK} methods due to it's intuitiveness and fast, but precise motion.

\item \textbf{\textit{H1.1.d.}} \textit{TB} will outperform other methods utilizing discrete motion (\textit{PH} and \textit{P}).

\end{itemize}

\begin{figure}[h]
    \centering
    \includegraphics[width=0.99\linewidth]{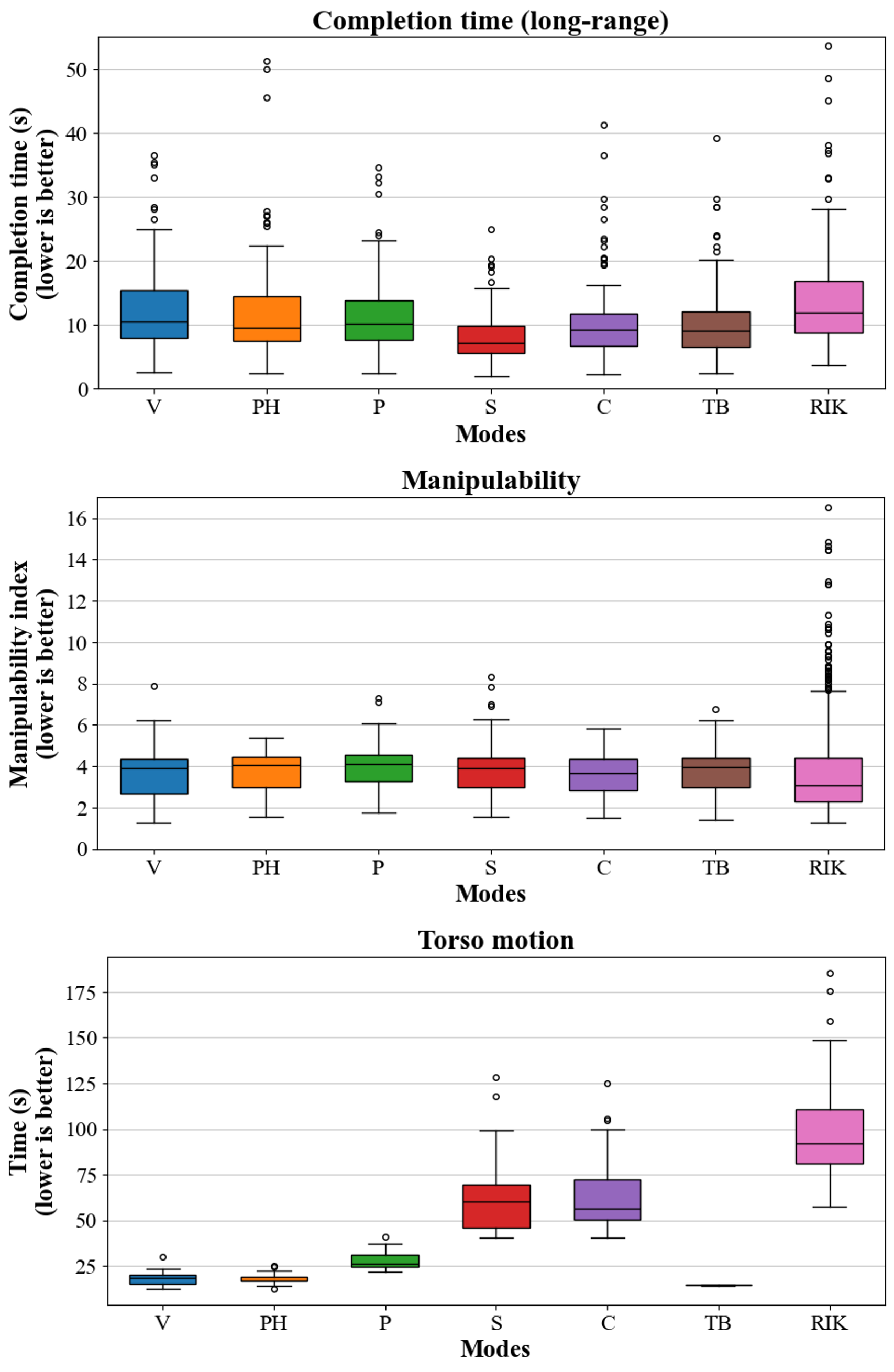}
    \caption{Results for \textbf{completion time} for long-range motion, \textbf{manipulability index} and \textbf{torso motion time} (metric of torso energy consumption). }
    \label{fig:robot}
    \vspace{-1ex}
\end{figure}

\textit{V} showed no significant difference in task completion time compared to other modes, leading to the rejection of hypothesis \textit{H1.1.a.} 
\textit{RIK} mode was significantly worse than all other methods, partially supporting hypotheses \textit{H1.1.b} and \textit{H1.1.c}. No significant difference was found between TB and (P, PH), hence, we rejected \textit{H1.1.d.}

\subsection{Short-range Completion Time}\label{sec:result-time}

\begin{itemize}
\item \textit{\textbf{H1.2.}} \textit{S} will be outperformed by other Reactive Autonomies (\textit{P} and \textit{C}) because of overshooting due to faster motion.
\end{itemize}

Only \textit{RIK} was significantly worse than all methods, leading to the rejection of \textit{H1.2}.

\subsection{Long-range Completion Time}\label{sec:result-time}

\begin{itemize}

\item \textit{\textbf{H1.3.a.}} \textit{PH} method will outperform \textit{V} method as a faster alternative to the baseline human-initiated method.

\item \textit{\textbf{H1.3.b.}} \textit{S} will outperform human-initiated methods (\textit{V} and \textit{PH}), all other Reactive Autonomies (\textit{P, C}) and \textit{RIK} method due to the scaled motion mapping.
\end{itemize}
    
\textit{S} $>$ (\textit{V, PH, P, C}), while \textit{C} $>$ (\textit{V, RIK}). \textit{TB} $>$ \textit{V}. This fully supports hypothesis \textit{H1.3.b} and rejects \textit{H1.3.a.}

\subsection{Manipulability Index}\label{sec:result-manipulability}

\begin{itemize}
\item \textit{\textbf{H2.} Arm's Manipulability}: \textit{S}, \textit{C} and \textit{RIK} will outperform all other methods.
\end{itemize}

\textit{RIK} was the best overall. 
\textit{C} $>$ \textit{S} $>$ (\textit{PH, P, TB}). \textit{V} $>$ (\textit{PH, P, S, TB}). This partially supports hypothesis \textit{H2}.

\subsection{Workload}\label{sec:result-workload}

\begin{itemize}
\item \textit{\textbf{H3.1.} Participant's Physical Workload}: \textit{S} will be outperformed by all Manual, other Reactive and Proactive Autonomy methods.

\item \textit{\textbf{H3.2.} Participant's Cognitive Workload}: Proactive Autonomy (\textit{TB}) will be the best, followed by Reactive Autonomies (\textit{P}, \textit{S} and \textit{C}) and \textit{RIK}, followed by \textit{Preset Height}, followed by \textit{V}.
\end{itemize}

No significant differences were found in cognitive or physical workload, leading to the rejection of hypotheses \textit{H3.1} and \textit{H3.2.}

\subsection{Torso Energy Consumption}\label{sec:result-energy}

\begin{itemize}
\item \textit{\textbf{H4.} Torso's Energy Consumption}: We hypothesize that methods that adjust the torso continuously will be outperformed by discrete methods. \textit{TB} will be the best, followed by (\textit{PH}, \textit{P}), followed by \textit{V}, followed by (\textit{S}, \textit{C}) and \textit{RIK} methods.
\end{itemize}

Significant differences were found in both torso motion time and torso motion fraction. (\textit{V, PH, P, TB}) $>$ (\textit{S, C, RIK}). Additionally, \textit{TB} $>$ \textit{P},  \textit{V} $>$ \textit{P}, \textit{PH} $>$ \textit{P}. These findings partially support hypothesis \textit{H4}.

\subsection{Subjective Data}\label{sec:result-subjective}

According to \textbf{NASA-TLX} \textit{S $>$ RIK} (\textit{p $<$ 0.05}). \textbf{SUS} revealed (\textit{V, PH, TB}) $>$ \textit{RIK} (\textit{p $<$ 0.05}), (\textit{S, C}) $>$ \textit{RIK} (\textit{p $<$ 0.001}). 

In our \textbf{custom survey} robotic arm control perception did not show any significant difference. However, significant differences were found in ease of torso control, with \textit{TB} $>$ \textit{RIK} \textit{(p $<$ 0.01)}. Similarly, alignment of torso motion with user intent showed that \textit{TB} $>$ \textit{RIK} (\textit{p $<$ 0.01}). In long-range motion efficiency, \textit{C} $>$ \textit{RIK} (\textit{p $<$ 0.05}), and \textit{TB} $>$ \textit{RIK} (\textit{p $<$ 0.01}). No significant differences were observed for short-range motions efficiency.

Participants also indicated their top three preferred modes for controlling the system. The most popular modes, selected based on user preference, were \textit{Chasing}, \textit{Scaling}, and \textit{Task-based}.

\section{Discussion and future work}\label{sec:con}

Several directions for future research emerge. Enhancing \textit{Task-Based} modes with gaze tracking, next-action probability models, and human intent estimation could improve predictive torso adjustments. Optimizing \textit{Preset Heights} settings based on specific task constraints, such as shelf levels, may enhance efficiency and intuitiveness. 

Without fine-tuning, \textit{RelaxedIK} prioritized torso motion. Since the torso moves significantly slower than the arm, this occasionally caused a mismatch where the arm reached the next goal while the torso lagged behind, prompting the operator to move the controller further and overshoot the target, resulting in an unpredictable system response.

In human-initiated methods (\textit{Velocity} and \textit{Preset Heights}), where operators had more freedom in selecting torso positions, our observations showed that no participants used simultaneous torso-arm adjustments. Instead, the majority followed a torso-first, arm-second strategy, aligning with a gross manipulation to precise alignment approach and a small group used the opposite. 

Although no significant differences in cognitive workload across modes were initially found, some individuals struggled more with certain modes, prompting further exploration via clustering. We performed K-means clustering based on math-solving rates. Although the number of participants per group is insufficient for definitive conclusions, we observed emerging patterns.

The analysis resulted in three clusters with normal workload distribution in each cluster: \textbf{Fast Cluster} (5 participants) favored dynamic modes like \textit{Chasing} and \textit{Scaling}, reflecting their quick decision-making and task-switching abilities, which led to faster math-solving speeds. \textbf{Moderate Cluster} (9 participants) preferred structured modes like \textit{Task-based}, balancing automation with manual control for reliable performance. \textbf{Slow Cluster} (3 participants) also leaned towards \textit{Task-based}, \textit{Chasing} and \textit{Scaling} but struggled more with human-initiated approaches (\textit{Velocity Control} and \textit{Preset Heights}), benefiting from guided, structured assistance.

No significant differences in physical workload were counterintuitive since methods like \textit{Scaling} clearly demand more effort. Since our estimation relied on upper-body posture tracking, we interpret this result as an effect of the Meta Quest Controller pose clutching (tracking) feature~\cite{boguslavskii2023shared}, which likely allowed participants to maintain more comfortable hand postures, effectively reducing measurable physical strain.

\section{Conclusion}\label{sec:con}

In this paper, we investigated various human-robot collaborative (HRC) control approaches for torso-arm coordination in mobile humanoid robots.
Results indicated that the Chasing and Scaling modes significantly improved task performance in long-range operations and were favored by participants alongside the Task-Based mode. These findings suggest that users preferred modes that offered a balance of dynamic adaptability and structured assistance, enhancing both efficiency and control in torso-arm coordination.
Our robot-initiated baseline, \textit{RelaxedIK}, underperformed and was the least favored because its torso-arm coordination often failed to match user intent. However, it was found to be the best for manipulability.
Although there were no significant differences in cognitive and physical workload across participants, preliminary clustering analysis identified three distinct groups whose preferences varied based on their levels of cognitive workload.
Future work will explore how users with different backgrounds and expertise levels perform across various HRC methods and task complexities to identify the most effective approaches for diverse operators.
\bibliographystyle{IEEEtran}
\bibliography{references}

\end{document}